\documentclass[10pt,twocolumn,letterpaper]{article}

\usepackage{iccv}
\usepackage{times}
\usepackage{epsfig}
\usepackage{graphicx}
\usepackage{amsmath}
\usepackage{amssymb}
\usepackage{dsfont}
\usepackage{array,multirow,graphicx}
\usepackage{algorithm}
\usepackage{cuted}
\usepackage{capt-of}
\usepackage[noend]{algpseudocode}
\def\etal{\textit{et al. }}

\usepackage{pdfrender}
\newcommand*{\boldcheckmark}{%
  \textpdfrender{
    TextRenderingMode=FillStroke,
    LineWidth=.5pt, 
  }{\checkmark}%
}

\usepackage[pagebackref=true,breaklinks=true,letterpaper=true,colorlinks,bookmarks=false]{hyperref}

\iccvfinalcopy 

\def\iccvPaperID{1485} 

\ificcvfinal\fi
\begin{document}

\title{Free-form Video Inpainting with 3D Gated Convolution and Temporal PatchGAN}
\author{Ya-Liang Chang$^*$ \and Zhe Yu Liu$^*$ \and Kuan-Ying Lee \and Winston Hsu \and \\
National Taiwan University, Taipei, Taiwan \\
{\tt\small \{yaliangchang, zhe2325138\}@cmlab.csie.ntu.edu.tw, \{r03922165, whsu\}@ntu.edu.tw}
}

\maketitle

\begin{strip}\centering
\vspace{-12mm}
\includegraphics[width=\linewidth]{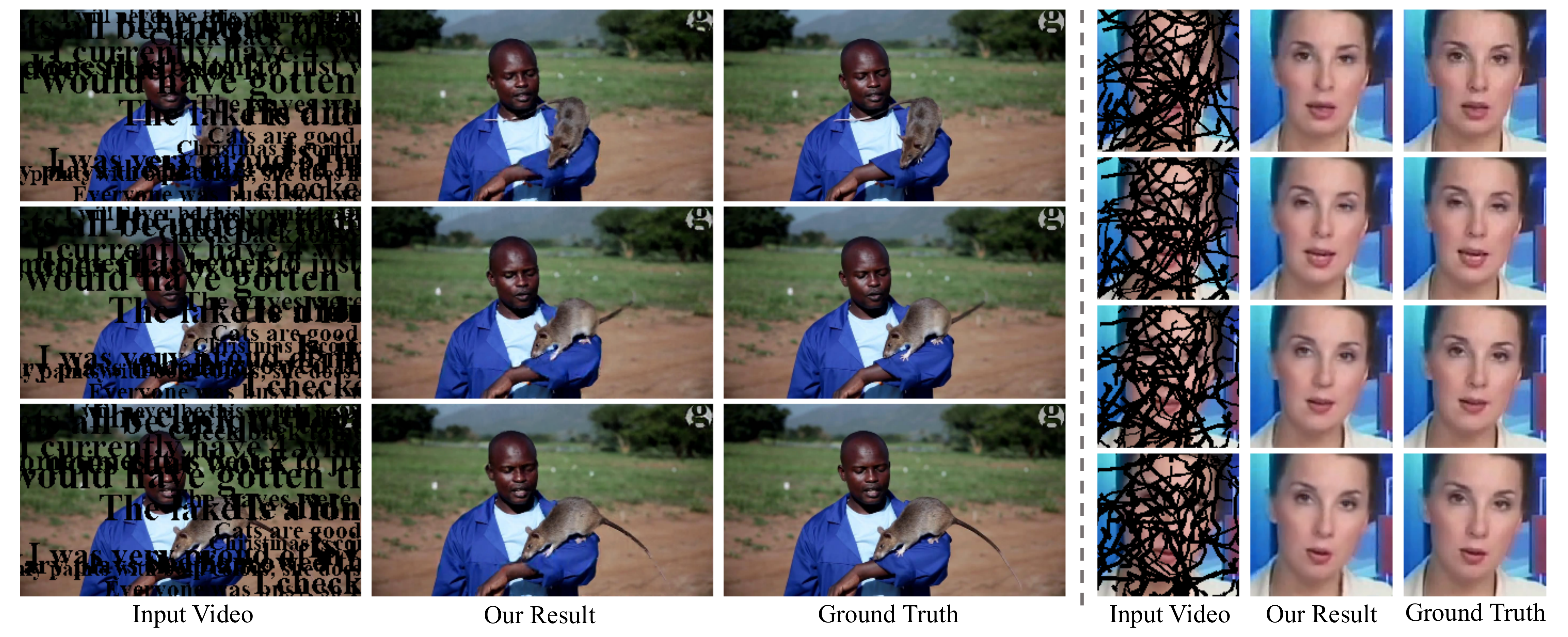}
\captionof{figure}{
Our model takes incomplete videsos with free-form masks (e.g. random text on the left and curves on the right) as inputs, and generates the completed videos as outputs. We propose using 3D gated convolutions to deal with the uncertainty of free-form masks and a combination of designed loss functions to enhance temporal consistency. Best viewed in \href{https://bit.ly/2Jz3lGg}{videos}.
\label{fig:teaser}}
\end{strip}
\renewcommand{\thefootnote}{\fnsymbol{footnote}}
\footnotetext[1]{The two authors contributed  equally to this paper.}
\begin{abstract}
\vspace{-1mm}
Free-form video inpainting is a very challenging task that could be widely used for video editing such as text removal (see Fig. \ref{fig:teaser}). Existing patch-based methods could not handle non-repetitive structures such as faces, while directly applying image-based inpainting models to videos will result in temporal inconsistency (see \href{http://bit.ly/2Fu1n6b}{videos}).
In this paper, we introduce a deep learning based free-form video inpainting model, with proposed 3D gated convolutions to tackle the uncertainty of free-form masks and a novel Temporal PatchGAN loss to enhance temporal consistency. In addition, we collect videos and design a free-form mask generation algorithm to build the free-form video inpainting (FVI) dataset for training and evaluation of video inpainting models. We demonstrate the benefits of these components and experiments on both the FaceForensics and our FVI dataset suggest that our method is superior to existing ones. Related source code, full-resolution result videos and the FVI dataset could be found on \href{https://github.com/amjltc295/Free-Form-Video-Inpainting}{Github}.
\vspace{-5mm}
\end{abstract}

\section{Introduction}
Video inpainting, to recover missing parts in a video, is a very challenging task that remains unsolved. It is a very practical and crucial problem and solving this problem could be beneficial for movie post-production and general video editing.
Among them, free-from video inpainting is the most difficult and unconstrained problem because the missing area could be of arbitrary shape (see Fig. \ref{fig:teaser}). In this paper, we propose a novel model to tackle the free-form video inpainting task, and both the quantitative and qualitative evaluations show our model can generate state-of-the-art results with high video quality.

There are many methods proposed for the video inpainting problem, such as patch-based algorithms \cite{granados2012not, Huang-SigAsia-2016, newson2014video, wexler2007space}, which aim to find the most similar patch from the unmasked parts of the video to fill in the masked region. However, patch-based models often fail to recover complex objects that could not be seen or found easily in the unmasked parts of the video. The nearest neighbor search algorithms in patch-based methods also may not work when the ratio of covered area by free-form masks to that of uncovered area is high (see Table \ref{tab:quantitative_comparison_VOR}). 

Aside from patch-based methods, many deep learning based models have made tremendous progress on free-form image inpainting. Nevertheless, simply applying image inpainting model to videos tends to cause twisted or flickering results that are temporally inconsistent (See \href{http://bit.ly/2FypRtV}{Edge-Connect}).

We extend the work of free-form image inpainting to videos by developing a novel architecture that enhances temporal consistency. The model is learning based, so it could model the data distribution based on the training videos and recover the masked regions. It could even recover objects that are mostly occluded in the video such as the face in Fig. \ref{fig:teaser}, which is impossible for patch-based methods.  Besides, our method fully utilizes the temporal information in videos, so the flickering problem of image inpainting is mitigated.

Specifically, we observe that input videos contain many masked voxels that are potentially harmful to vanilla convolutions, we design a generator with 3D gated convolutional layers that could properly handle the masked video by learning the difference between the unmasked region, filled in region and masked region in each layer and attend on proper features correspondingly. In addition, different from image inpainting, video inpainting has to be temporally coherent, so we propose a novel Temporal PatchGAN discriminator that penalizes high-frequency spatial-temporal features and enhances the temporal consistency through the combination of different losses. We also design a new algorithm to generate diverse free-form video masks, and collect a new video dataset based on existing videos that could be used to train and evaluate learning-based video inpainting models.

Our contributions could be summarized as follows:
\begin{itemize}
\item We extend the work of image inpainting and propose the first learning-based model for free-form video inpainting and achieve state-of-the-art results qualitatively and quantitatively on the FaceForensics and our dataset.
\item We introduce a novel Temporal PatchGAN (T-PatchGAN) discriminator to enhance the temporal consistency and video quality. It could also be extended to other video generation tasks such as video object removal or video super-resolution.
\item We design a new algorithm to generate free-form masks. We design and evaluate several types of masks with different mask-to-frame ratios.
\item We collect the free-form video inpainting (FVI) dataset, the first dataset to our knowledge for training and evaluation of free-form video inpainting methods, including 1940 videos from the YouTube-VOS \cite{xu2018youtube} dataset and 12600 videos from the YouTube-BoundingBoxes \cite{real2017youtube} dataset.

\end{itemize}

\section{Related Work}

\paragraph{\textbf{Image Inpainting.}}
Image inpainting, to recover the damaged or missing region in a picture, is firstly introduced in \cite{bertalmio2000image}. Many approaches have been proposed to solve the image inpainting task, including diffusion-based \cite{bertalmio2001navier, bertalmio2000image} and patch-based \cite{barnes2009patchmatch, bornard2002missing, drori2003fragment} ones. In general, these methods performs well on simple structure but often fails to generate complex objects or recover large missing area.

Over the past few years, deep learning based methods have made tremendous progress on image inpainting.  Xie \etal \cite{xie2012image} is the first to introduce convolutional neural networks (CNNs) for image inpainting and denoising on small regions. Subsequently, Pathak \etal \cite{pathak2016context} extended image inpainting to larger region with an encoder-decoder structure and used generative adversarial network (GAN) \cite{goodfellow2014generative} where a generator that strives to create genuine images and a discriminator learns to recognize fake ones are jointly trained to improve the blurry issue caused by the $l_2$ loss. Yu \etal \cite{yu2018generative} further proposed a contextual attention layer with local and global WGANs to achieve better results.

\paragraph{\textbf{Free-form Image Inpainting.}}
Recently, image inpainting with irregular holes (free-form masks) caught more attention because it is closer to the real case. Yan \etal \cite{yan2018shift} designed a special shift-connection layer in the U-Net architecture; Lui \etal \cite{liu2018image} proposed the partial convolution; Yu \etal \cite{yu2018free} developed the gated convolution with spectral-normalized discriminator to improve free-form image inpainting. Asides from these works, Nazeri \etal \cite{nazeri2019edgeconnect} proposed a two-stage adversarial model EdgeConnect, where the edge generator firstly hallucinates edges of the missing region, and the image completion network generates the final output image using hallucinated edges as a priori. Nazeri \etal provided a pretrained model that reaches state-of-the-art, which we set as a baseline in our work.

Although state-of-the-art image inpainting models could recover missing regions in a picture in a reasonable manner, extending these models to videos will cause serious temporal inconsistency as each inpainted frame is different (see \href{http://bit.ly/2UPyaYp}{videos}).

\begin{figure*} [ht]
\begin{center}
\includegraphics[width=480pt]{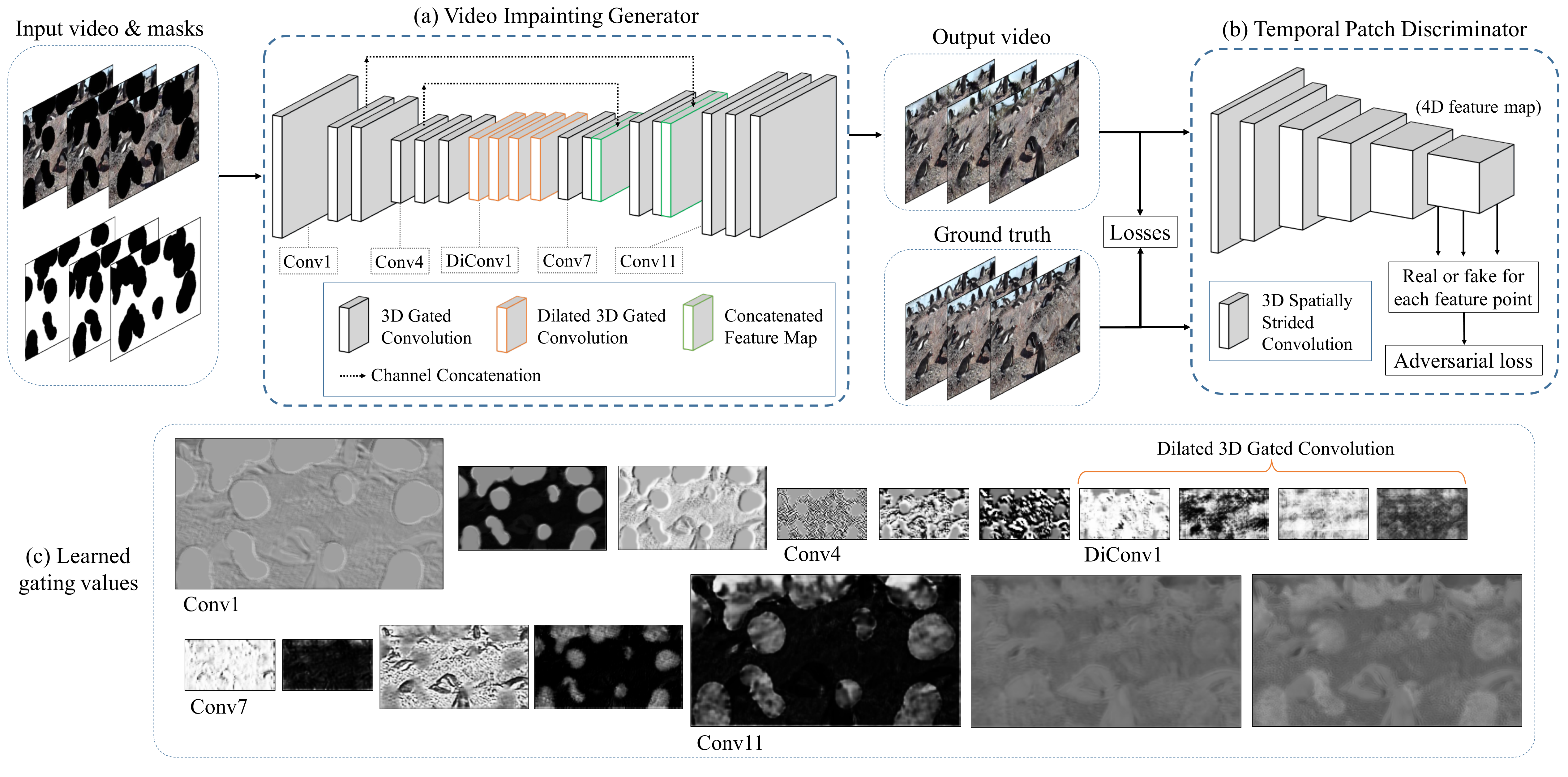}
\vspace{-2mm}
\end{center}
\caption{Model architecture and learned gated value visualization. Our model is composed of (a) a video inpainting generator with 3D gated convolutional layers that fully utilizes information for neighboring frames to handle irregular video masks and (b) a Temporal PatchGAN (T-PatchGAN) discriminator that focuses on different spatial-temporal features to enhance output quality. (c) The visualization of learned gating values $\sigma(Gating_{t,x,y})$. The 3D gated convolution will attend on the masked area and gradually fill in the missing feature points.
Note that all gating values are extracted from the first channel of each layer without manual picking.}
\label{fig:model_architecture}
\vspace{-2mm}
\end{figure*}

\paragraph{\textbf{Video Inpainting.}}
Traditionally, patch-based methods \cite{granados2012not, Huang-SigAsia-2016, newson2014video, wexler2007space} are used for video inpainting. 
Wexler \etal \cite{wexler2007space} considered video inpainting as a global optimization problem, where all missing regions could be filled in with similar patches from the unmasked parts. 
Afterwards, Newson \etal \cite{newson2014video} further improved the search algorithm, integrating texture features and initialization scheme. 
Lastly, Huang \etal \cite{Huang-SigAsia-2016} tackled the moving camera problem by jointly estimating optical flow and colors in the missing regions, and we also take it as our baseline. 

State-of-the-art patch-based methods could generate plausible videos under certain conditions, but the computation time of these methods is too high for real-time applications. 
In addition, patch-based models are limited to repetitive patterns or appeared objects and not feasible to complex structures and large/long-lasting occlusions. 
The proposed model is learning based that could solve both problems by modeling the distribution of real videos and generate realistic results only by forward inference, without searching.

To solve issues in patch based methods, Wang \etal \cite{wang2018videoinp} proposed the first deep learning based method CombCN for video inpainting, and we also set it as a baseline. It is a two-stage model with a 3D convolutional network for temporal consistency, followed by a 2D completion network to improve video quality. Although their model could be applied to some random holes during validation/testing stage, it is rather limited so we do not consider it as a free-form video inpainting method.
Besides, their model only uses traditional convolution and is trained with the $l_1$ loss, so the results tend to be blurry in complex scenes \cite{chang2019vornet}. Our model is single-stage, uses gated convolution to attend on valid features, and integrates perceptual and temporal generative adversarial loss to generate clear and plausible videos for irregular moving masks.

\section{Proposed Method}
The proposed model (Fig. \ref{fig:model_architecture}) consists of a generator network $G$ with 3D gated convolution to inpaint videos, and a Temporal PatchGAN discriminator $D$ with several losses.

\subsection{Video Inpainting Generator}
We extend the single-stage UNet-like network used for image inpainting \cite{liu2018image} to video inpainting and integrate the gated convolutional layers in \cite{yu2018free} (see Fig. \ref{fig:model_architecture} (a)). During training, we combine ground truth video frames $\{V_t  \mid  t=1 \dots n\}$ and masks $\{M_t  \mid  t=1 \dots n\}$ into masked input video $\{I_t  \mid  t=1 \dots n\}$. The model will inpaint the masked region and generate the output video frames $\{O_t  \mid  t=1 \dots n\}$.

\subsection{Spatial-temporally Aware 3D Gated Conv.}
In vanilla convolutional layers, all pixels are treated as valid ones, which makes sense for tasks with real images/videos as inputs, such as object detection or action recognition. However, for inpainting problems, masked regions are filled with black pixels, so input features for convolutional layers include invalid pixels (shallow layers) or synthesized pixels (deep layers), which should not be treated exactly as normal ones. 

To address this problem, we propose the 3D gated convolutions extended from \cite{yu2018free} to replace vanilla convolutions in our generator. The 3D convolutions utilizes information from neighboring frames, while the gated convolutions attends on the irregular masked areas; together, 3D gated convolutions could properly handle the uncertainty of free-form video masks.  Specifically, for each convolutional layer, an additional gating convolutional filter $W_g$ is applied to input features $F_{t, x, y}$ to obtain a gating $Gating_{t, x, y}$ , which is used as an attention map on the output features $Features_{t, x, y}$ from original vanilla convolutional filter $W_f$ according to the validity (see Fig. \ref{fig:model_architecture}(c)). The $t, x, y$ are the spatial-temporal coordinates of the video. It could be expressed as:
\vspace{-2mm}
\begin{equation}
Gating_{t, x, y} = \sum \sum W_g \cdot F_{t, x, y} 
\end{equation}
\vspace{-3mm}
\begin{equation}
Features_{t, x, y} = \sum \sum W_f \cdot F_{t, x, y} 
\end{equation}
\begin{equation}
Output_{t, x, y} = \sigma(Gating_{t, x, y})  \phi(Features_{t, x, y})
\end{equation}
where $\sigma$ is the sigmoid function to transform gating to values between 0 (invalid) and 1 (valid), and $\phi$ is the original activation function (e.g. LeakyReLU).

\subsection{Loss Functions}
The overall loss function to train the model is defined as:
\vspace{-1mm}
\begin{equation}
\begin{aligned}
L_{total} &= \lambda_{l_1} L_{l_1}    +
\lambda_{{l_1}_{mask}}  L_{{l_1}_{mask}}    +
\lambda_{perc}  L_{perc}   \\      &+
\lambda_{style} L_{style} + 
\lambda_{G} L_{G}
\end{aligned}
\end{equation}
where $\lambda_{l_1}$, $\lambda_{{l_1}_{mask}}$, $\lambda_{perc}$, $\lambda_{style}$ and $\lambda_{G}$ are the weights for $l_1$ loss, masked $l_1$ loss, perceptual loss, style loss and Temporal PatchGAN loss, respectively. 

\vspace{-3mm}
\paragraph{\textbf{Masked $l_1$ loss.}} The $l_1$ loss focuses on the pixel-level features. Since the unmasked area will be pasted onto the final output video, we separate the $l_1$ loss for all videos:
\begin{equation}
L_{l_1} = \mathds{E}_{t,x,y}[ |O_{t,x,y} - V_{t, x, y}|]
\end{equation}
and the $l_1$ loss for the masked region:
\begin{equation}
L_{{l_1}_{mask}} = \mathds{E}_{t,x,y}[ M_{t,x,y} |O_{t,x,y} - V_{t, x, y}|]
\end{equation}

\paragraph{\textbf{Perceptual loss.}} Perceptual loss is firstly proposed in \cite{gatys2015neural} to keep image contents for style transfer, and is now widely used for image inpainting \cite{liu2018image, nazeri2019edgeconnect} and super-resolution \cite{johnson2016perceptual, ledig2017photo} to mitigate the blurriness caused by the $l_1$ loss. The perceptual loss computes the $l_1$ loss in feature level:
\vspace{-2mm}
\begin{equation} \label{equ6}
L_{perc} = \sum_{t=1}^{n} \sum_{p=0}^{P-1} \frac{|\Psi^{O_{t}}_p - \Psi^{V_{t}}_p|}{N_{\Psi^{V_{t}}_p}}
\vspace{-2mm}
\end{equation}
where $\Psi^{V_{t}}_p$ denotes the activation from the $p$th selected layer of the pretrained  network given the input $V_t$, and $N_{\Psi^{V_{t}}_p}$ is the number of elements in the $p$th layer. We use layer $relu_{2\_2}$, $relu_{3\_3}$ and $relu{4\_3}$ from the VGG \cite{simonyan2014very} network pre-trained on ImageNet \cite{russakovsky2015imagenet}.

\vspace{-4mm}
\paragraph{\textbf{Style loss.}} We also include the style loss, which is introduced in \cite{gatys2015neural} to keep the image style for style transfer, and also used in image inpainting \cite{liu2018image, nazeri2019edgeconnect}. Style loss is similar to perceptual loss, except that an auto-correlation (Gram matrix) is firstly applied to the features:
\begin{equation}
L_{style} = \sum_{t=1}^{n} \sum_{p=0}^{P-1} \frac{1}{C_p C_p} \frac{|(\Psi^{{O_t}_p})^T(\Psi^{{O_t}_p}) - (\Psi^{V_{t}}_p)^T(\Psi^{V_{t}}_p))|}{C_p H_p W_p}
\end{equation}
where $\Psi^{{O_t}_p}$ and $\Psi^{{V_t}_p}$ are both VGG features in the shape of ($H_p$, $W_p$, $C_p$) as the ones in perceptual loss \ref{equ6}.

\vspace{-3mm}
\paragraph{\textbf{Temporal PatchGAN loss.}} 

For the free-form video inpainting problem, masks could be anywhere in a video, so we should consider global and local features in each frame, and the temporal consistency of these features. A naive idea will be applying a loss function for each of the three aspects respectively. However, empirically we found that it is hard to balance the weights of these loss functions, especially when some of them are GAN losses (adding GAN loss is a very common strategy to make image inpainting results more realistic \cite{nazeri2019edgeconnect, yu2018generative, yu2018free}). 

Yu \etal proposed an efficient SN-PatchGAN \cite{yu2018free}, which applies GAN loss on feature maps of the discriminator to replace the use of global and local GAN and tackle the problem that masks could be anywhere and of any form. Although their work tackles the balancing issue between GAN losses and solves free-form image inpainting problem, it does not consider temporal consistency, a pivotal factor for high-quality video inpainting. Inspired by their work, we further integrate the temporal dimension and design a novel Temporal PathGAN (T-PatchGAN) discriminator that focuses on different spatial-temporal features to fully utilize all the global and local image features and temporal information together. 

Our T-PatchGAN discriminator is composed of 6 3D convolutional layers with kernel size $3 \times 5 \times 5$ and stride $1 \times 2 \times 2$. The recently proposed spectral normalization \cite{miyato2018spectral} is applied to both the generator and discriminator, similar to \cite{nazeri2019edgeconnect} to enhance training stability. In addition, we use the hinge loss as the objective function as to discriminate if the input video is real or fake:
\begin{equation}
\begin{aligned}
L_{D} &= \mathds{E}_{x\sim P_{data}(x)}[ReLU(1+D(x))] \\ &+
\mathds{E}_{z\sim P_{z}(z)}[ReLU(1-D(G(z)))]
\end{aligned}
\end{equation}

\begin{equation}
\begin{aligned}
L_{G} = -\mathds{E}_{z\sim P_{z}(z)}[D(G(z))]
\end{aligned}
\end{equation}
where G is the video inpainting network that takes input video z and D is the T-PatchGAN discriminator. 

Note that we use kernel size $3 \times 5 \times 5$ for each layer in the discriminator, so the receptive field of each output feature covers the whole videos, and a global discriminator like \cite{yu2018generative} is not needed. The T-PatchGAN learns to classify each spatial-temporal patch as real or fake, which restricts it to focus on high-frequency features because it only penalizes at the scale of patches. As the $l_1$ loss already focus on low-frequency features, using T-PatchGAN could improve the output video quality in an efficient way.

\subsection{Free-form Video Masks Generation}

\begin{figure} [ht]
\begin{center}
\includegraphics[width=210pt]{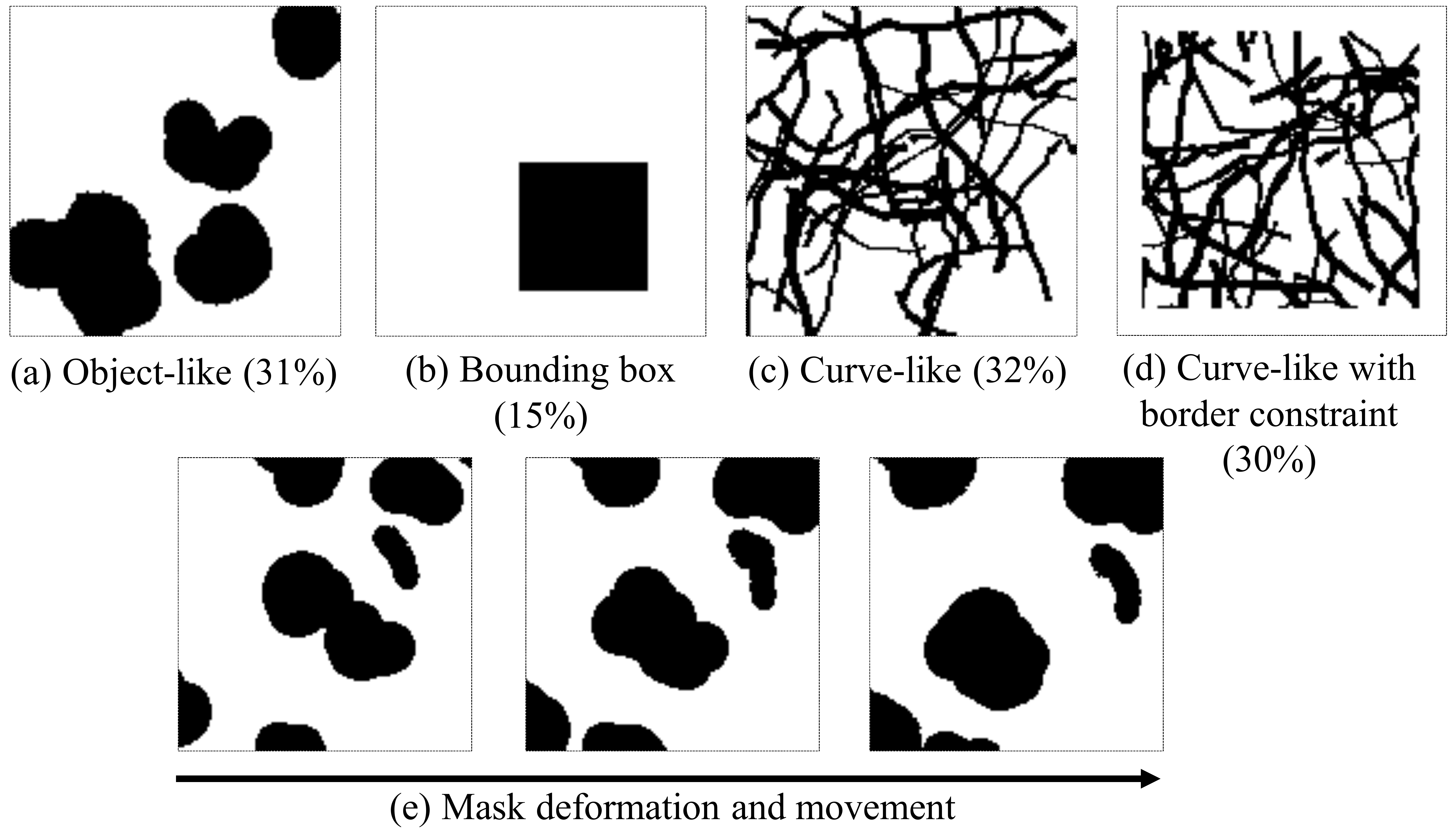}

\end{center}
\caption{Masks generated by our algorithm with different mask types and mask-to-image ratios. The components in a mask video may move and deform independently as shown in (e).}
\label{fig:mask_explanation}
\vspace{-4mm}
\end{figure}

Training data is extremely important for learning based methods, and the generation of our input mask videos is non-trivial as it should consider different scenarios to be ``free-form''. There is no existing database or algorithm to generate such free-form video masks, so we develop an video mask generation algorithm based on the image one by \cite{yu2018free} (see Algorithm 1 in the supplementary materials).

The image mask generation in \cite{yu2018free} uses several strokes drawing on a blank image to represent a mask. Each stroke is composed of a ordered set of control points, which is determined by the trace of a head point initialized at a random position and repeatedly moving to a nearby position. 



Additionally,  for free-form video masks, we introduce the concept of motion: strokes may move and deform over time (see Fig. \ref{fig:mask_explanation}(e)). Stroke deformation is achieved by randomly moving each control point of a stroke with a certain probability. For the movement, the concepts of velocity and acceleration are applied on the strokes.
The initial speeds of strokes are sampled from a normal distribution centered on 0 since most objects in videos do not have large speeds.

As stated in \cite{liu2018image}, many methods \cite{iizuka2017globally, liu2018image} have degraded performance when masks cover the border. Therefore, aside from motion simulation, we also take such border constraints into consideration. That is, we generate both masks that either cover or do not cover the edges of the frame (see Fig. \ref{fig:mask_explanation}. Masks without border constraint are more difficult since convolutional filters will have no valid pixels as inputs in the masked border areas. 

Moreover, we consider three different types of masks: long thin curve-like and round thick object-like masks generated with different hyper-parameters in our mask generation algorithm along with bounding-box masks, as shown in Fig. \ref{fig:mask_explanation}. The curve-like masks are considered easier as most masked areas are close to valid pixels (unmasked area), while object-like and bounding-box masks are challenging since it is hard to inpaint large invalid voxels. 

Totally 28,000 free-form videos with mask-to-frame ratio from 0 - 10\% to 60 - 70\% are generated for training. And for each mask type, 100 videos are generated for testing.

\section{Experimental Results}
\subsection{Datasets}
\paragraph{\textbf{FaceForensics.}} We compare with \cite{wang2018videoinp} on the FaceForensics dataset \cite{rossler2018faceforensics}, which contains 1004 face videos from YouTube and the YouTube-8m dataset \cite{abu2016youtube} with ”face”, ”newscaster” or ”newsprogram” tags. The videos are cropped into 128 $\times$ 128 with the face in the middle during the data preparation stage, following the setting in \cite{wang2018videoinp}. Among them, 150 videos are used for evaluation and the rest are used for training. The FaceForensics dataset is rather easy for learning-based methods since the data is less diverse.

\vspace{-3mm}
\paragraph{\textbf{Free-form video inpainting (FVI) dataset}} To test on more practical cases, we collect videos from the YouTube-VOS \cite{xu2018youtube} dataset and the YouTube-BoundingBoxes dataset \cite{real2017youtube}. The former has about 2000 videos with 94 categories of object segmentation in 6 frame per second (FPS) and the latter has about 380,000 videos with 23 kinds of object bounding boxes in 30 FPS.  We choose videos with resolution higher than 640 $\times$ 480 and manually filter out videos with shot transitions. We set 100 videos from the YouTube-VOS as testing set, while the training set includes about 15,000 videos.  Together with the 28,000 free-form mask videos, we build the FVI dataset, the first dataset for free-form video inpainting, to the best of our knowledge.

Our FVI dataset is very challenging for the video inpainting task due to the high diversity, including different kinds of objects, animals and human activities. All videos are from YouTube, closer to real-world scenario. Moreover, the provided object segmentation and bounding boxes could be used to test video object removal.

For the experiments, we only use 1940 videos from the training set as we do not witness a significant improvement for our model using the full training set. During data pre-processing stage, we resize videos to 384 $\times$ 216 and randomly crop them to 320 $\times$ 180 with random horizontal flip. 



\subsection{Evaluation metrics} We use mean square error (MSE) 
and Learned Perceptual Image Patch Similarity (LPIPS) \cite{zhang2018unreasonable} to evaluate the image quality. 
Furthermore, to evaluate the video quality and temporal consistency, we also calculate the Fréchet Inception Distance (FID) \cite{heusel2017gans} with I3D \cite{carreira2017quo} pre-trained video recognition CNN as Vid2vid \cite{wang2018video}.
See the supplementary materials for the details.

\subsection{Quantitative Results}
\label{sec:qal}
\begin{figure} [H]
\begin{center}
\includegraphics[width=\linewidth]{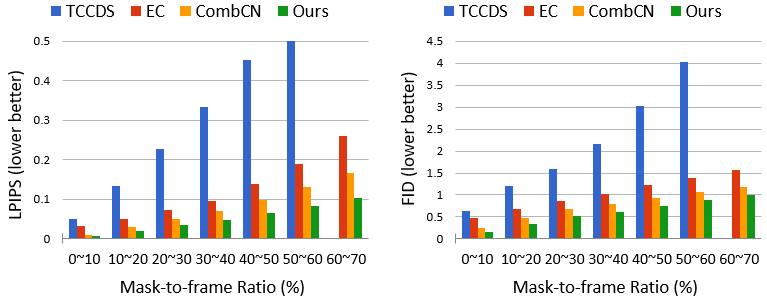}
\end{center}
\vspace{-4mm}
\caption{Effect of mask size on LPIPS and FID on the FVI test set with curve-like masks. Larger masks are harder for all methods.}
\label{fig:VOR_rand_curve_differnt_mask_ratio}
\vspace{-3mm}
\end{figure}

We evaluate our model on the FaceForensics and FVI testing set with free-form masks in 7 ranges of mask-to-frame ratio from  0 - 10\% to 60 - 70\% (higher mask-to-frame ratio makes the task more difficult, see Fig. \ref{fig:VOR_rand_curve_differnt_mask_ratio}). The state-of-the-art patch-based video inpainting method TCCDS by Huang \etal \cite{Huang-SigAsia-2016}, image inpainting method Edge-Connect(EC) by Nazeri \etal \cite{nazeri2019edgeconnect} and learning based video inpainting method CombCN by Wang \etal \cite{wang2018videoinp} are set for comparison. We train Nazeri \etal's model and Wang \etal's model on our dataset. Note that Wang \etal's model is originally trained on bounding boxes, but we train it with our free-form mask for fair comparison.

From Table \ref{tab:quantitative_comparison_FaceForensics} we could see that the FaceForensics dataset is easy for learning-based models but not for the patch-based method TCCDS \cite{Huang-SigAsia-2016}, because face features are non-repetitive and hence cannot be recovered with nearby patches. Yet, the overall structure of faces are learnable and thus learning-based methods are favorable. Compared with the two deep learning based methods, our model has superior performance on curve-like and object-like masks since it fully utilizes information of neighboring frames to recover the missing areas by 3D convolutions and the proposed T-PatchGAN loss. As for bounding-box masks, our model outperforms CombCN while on par with EC. Note our model is only trained on FVI. Hence, for fair comparison, we train EC from scratch on FVI without having it pre-trained on Celeb-A as stated in the original paper.

On the other hand, Table \ref{tab:quantitative_comparison_VOR} shows that our FVI dataset is more challenging for learning-based methods for its high diversity. It is rather difficult for learning-based models to capture the distribution of the highly diverse masks, while patch-based methods like TCCDS could easily find realistic enough patches to fill in the mask given the mask is not large. Nonetheless, we could notice that for some masks, the nearest neighbor search in TCCDS fails to find candidates when most patches are covered by the mask. Note that CombCN is only trained with the $l_1$ loss, so while it reports a lower MSE, its results are actually blurry, bearing high perceptual distance to ground truths (high LPIPS). Our method generates clear results (low LPIPS and FID) and demonstrates high temporal consistency (low FID), which is of crucial importance in video inpainting task.

\subsection{Qualitative Results}
\begin{figure} [tp]
\begin{center}
\includegraphics[width=240pt]{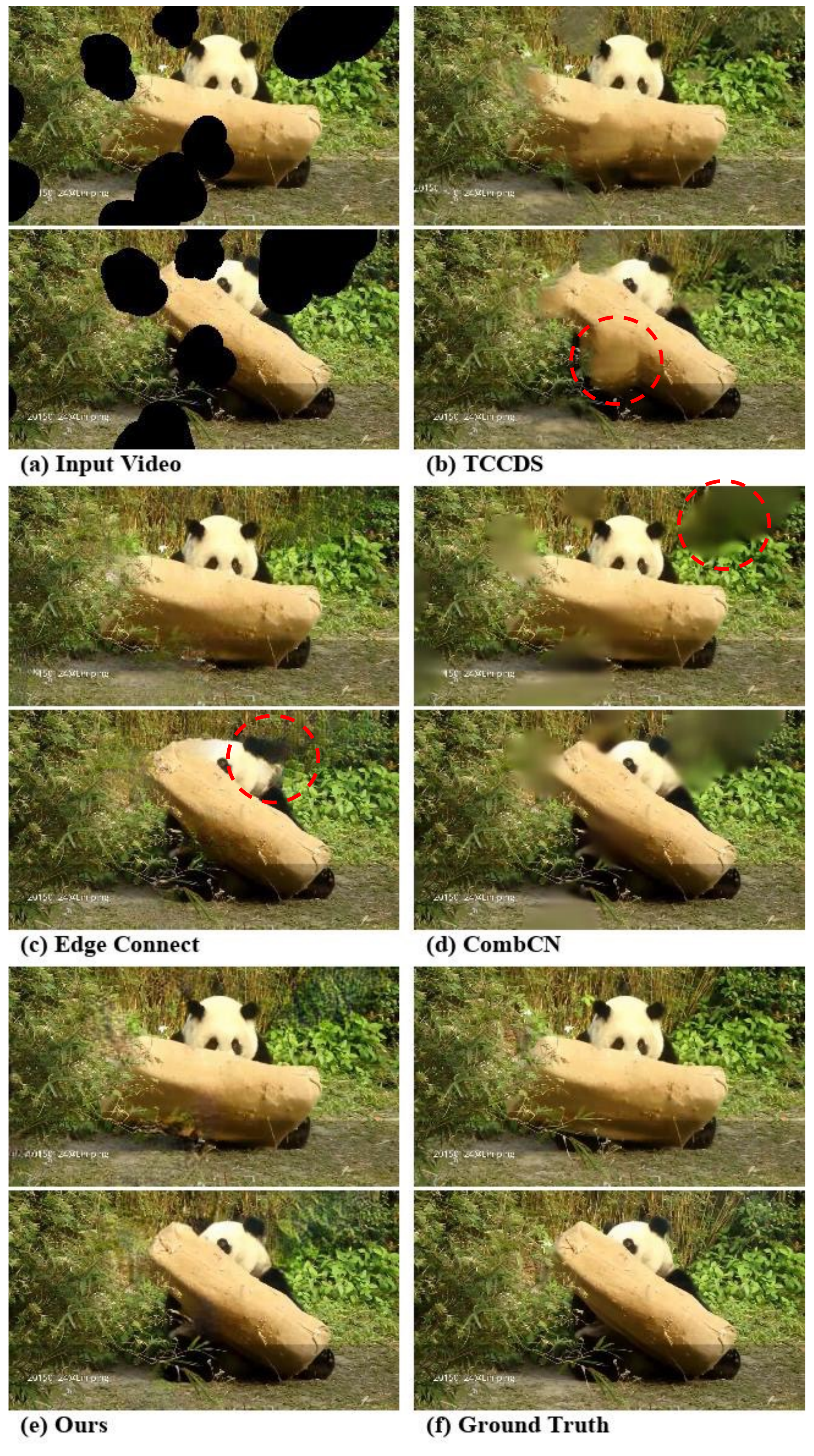}
\end{center}
\vspace{-3mm}
\caption{Visual comparison with the baselines. TCCDS: pasting wrong patches; Edge-Connect: inconsistent between frames; CombCN: blurry. See \href{http://bit.ly/2Fu1n6b}{vidoes}}
\vspace{-3mm}
\label{fig:visual_comparison}

\end{figure}

We also demonstrate the visual comparison in Fig. \ref{fig:visual_comparison} with the corresponding video link. As mentioned in \ref{sec:qal}, CombCN's \cite{wang2018videoinp} outputs are blurry due to the $l_1$ loss, TCCDS \cite{Huang-SigAsia-2016} may paste wrong patches, and Edge-Connect \cite{nazeri2019edgeconnect} will have flickering results (best viewed in \href{http://bit.ly/2FwbZl4}{videos}). 
Our model could generate reasonable frames with high temporal consistency. 

In addition, our trained model can be easily applied on object removal, as shown in Fig. \ref{fig:object_removal_examples}. More visual comparisons could be found in the supplementary materials. 

 \begin{table}[H]

 \centering
 \begin{tabular}{|c|c||c|c|c|c|}
 \hline
  \begin{tabular}[c]{@{}c@{}}\end{tabular} &
 \begin{tabular}[c]{@{}c@{}}Mask \\ Type\end{tabular} &
 \begin{tabular}[c]{@{}c@{}}TCCDS \end{tabular} &
 \begin{tabular}[c]{@{}c@{}}EC \end{tabular} &
 \begin{tabular}[c]{@{}c@{}}CombCN \end{tabular} &
 \begin{tabular}[c]{@{}c@{}}3DGated \\ (Ours) \end{tabular} \\
  \hline  \hline 	
 \parbox[t]{2mm}{\multirow{3}{*}{\rotatebox[origin=c]{90}{MSE$\downarrow$}}} 
& Curve & 0.0031*  & 0.0022  &  0.0012 &  \textbf{0.0008}  \\
& Object &  0.0096* &  0.0074 &  \textbf{0.0047} &  0.0048   \\
& BBox &  0.0055 &  0.0019 &  \textbf{0.0016} &  0.0018   \\


 \hline \hline 
 \parbox[t]{2mm}{\multirow{3}{*}{\rotatebox[origin=c]{90}{LPIPS$\downarrow$}}}
& Curve &  0.0566* &  0.0562 &  0.0483 &  \textbf{0.0276}  \\
& Object &  0.1340* &  0.0761 &  0.1353 &  \textbf{0.0743}   \\
& BBox &  0.1260 &  \textbf{0.0335} &  0.0708 &  0.0395   \\

 \hline \hline  
 \parbox[t]{2mm}{\multirow{3}{*}{\rotatebox[origin=c]{90}{FID$\downarrow$}}} 
& Curve & 1.281*  & 0.848  &  0.704 &  \textbf{0.472}  \\
& Object &  1.107* & 0.946  &  0.913 &  \textbf{0.766}   \\
& BBox &  1.013 &  \textbf{0.663} &  0.742 &  \textbf{0.663}   \\
 \hline
 \end{tabular}
 \caption{Quantitative results on the FaceForensics testing set with masks without border. 
 Our model has superior performance for the curve-like and object-like masks.
 *TCCDS fails on some masks; the results are average of the successful cases.
}
\label{tab:quantitative_comparison_FaceForensics}
\vspace{-5mm}
\end{table}

 \begin{table}[H]
 \centering

  \begin{tabular}{|c|c||c|c|c|c|}
 \hline
  \begin{tabular}[c]{@{}c@{}}\end{tabular} &
 \begin{tabular}[c]{@{}c@{}}Mask \\ Type\end{tabular} &
 \begin{tabular}[c]{@{}c@{}}TCCDS \end{tabular} &
 \begin{tabular}[c]{@{}c@{}}EC$^\#$ \end{tabular} &
\begin{tabular}[c]{@{}c@{}}CombCN \end{tabular} &
    \begin{tabular}[c]{@{}c@{}}3DGated \\ (Ours) \end{tabular} \\
  \hline  \hline 
 \parbox[t]{2mm}{\multirow{2}{*}{\rotatebox[origin=c]{90}{MSE}}} 
& Curve & 0.0219* &	0.0047 &	\textbf{0.0021} &	0.0024  \\
& Object & 0.0110* &	0.0079 &	\textbf{0.0049} &	0.0056   \\ 

 \hline \hline 
 \parbox[t]{2mm}{\multirow{2}{*}{\rotatebox[origin=c]{90}{LPI.}}}
& Curve &  0.2838* &	0.1204 &	0.0794 &	\textbf{0.0521}  \\
& Object & 0.2001* &	0.1420 &	0.2054 &	\textbf{0.1078}   \\

 \hline \hline  
 \parbox[t]{2mm}{\multirow{2}{*}{\rotatebox[origin=c]{90}{FID}}} 
& Curve  &   2.105* &	1.033 &	0.766 &	\textbf{0.609} \\
& Object & 1.287* &	1.083 &	1.091 &	\textbf{0.905} \\ 

 \hline
 \end{tabular}
 \caption{Quantitative comparison 
 on the FVI testing set without border. The results are the average of different mask-to-frame ratios (see the supplementary materials for original data).
 Our model outperforms the baselines for perceptual distance (LPIPS) and temporal consistency (FID). CombCN has better MSE but their results are blurry (see Fig. \ref{fig:visual_comparison}). 
 $^\#$Pretrained on Places2 \cite{zhou2017places}. 
}
\label{tab:quantitative_comparison_VOR}
\vspace{-4mm}
 \end{table}

\subsection{User Study}
Aside from qualitative comparison, we also conduct a human subjective study to evaluate our method. During the study, we display a pair of result videos  (ours against baselines or ground truth, in random sequence), and ask subjects to choose the more realistic and consistent one. The mask video is shown meanwhile for reference. For each mask type (object-like and curve-like) and mask-to-hole ratio (0-10\% to 60-70\%), we randomly select 20 video pairs to compare and each video pair is presented to 5 subjects.

\begin{figure} [tp]
\begin{center}
\includegraphics[width=\linewidth]{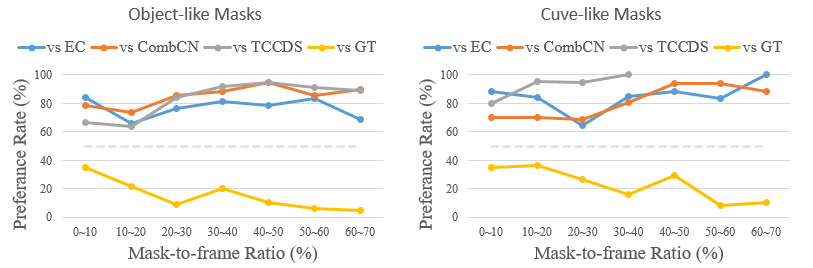}
\end{center}
\vspace{-3mm}
\caption{User preference on the FVI testing set (ours versus baselines and ground truth). 50\% means that the two methods are equally good. Our model outperforms the baselines in both the object-like and curve-like masks for all mask-to-frame ratios. When compared with ground truth (GT), our method could still have about 23\% preference for curve-like masks. }
\label{fig:user_study}
\vspace{-4mm}
\end{figure}

Results from 150 participants are shown in Fig. \ref{fig:user_study}. Our model outperforms the baselines in both object-/curve-like masks for all mask-to-frame ratios. In addition, when compared with ground truth, our method still has 23\% preference on average, which indicates subjects could not tell our results and the original videos apart in 23\% cases.

\subsection{Ablation Study}
We conduct an ablation study to evaluate the contribution of each proposed component. From Table \ref{tab:ablation_study} we could see that 3D convolution and T-PatchGAN are both crucial because the two components provide a great amount of temporal information by 3D convolutions. Corresponding video comparisons could be found on \href{https://bit.ly/2Fwbibs}{YouTube}.

 \begin{table}[H]
 \begin{center}
\begin{tabular}{|ccc|cc|}
\hline
\begin{tabular}[c]{@{}c@{}}3D \\ conv.\end{tabular} &
\begin{tabular}[c]{@{}c@{}}Gated \\ conv. \end{tabular} & 
\begin{tabular}[c]{@{}c@{}c@{}}T-Patch \\ GAN  \end{tabular} & 
LPIPS$\downarrow$  & FID $\downarrow$ \\ \hline
 &  $\boldcheckmark$ &   $\boldcheckmark$    &  0.1769  &  1.243  \\
$\boldcheckmark$  &   &   $\boldcheckmark$  &   0.1321   &    1.121 \\
$\boldcheckmark$  &  $\boldcheckmark$ &         &   0.1716  &   1.201    \\
$\boldcheckmark$  & $\boldcheckmark$  & $\boldcheckmark$    &  \textbf{0.1209}     & \textbf{1.034} \\
 \hline
\end{tabular}
\caption{Ablation study on the FVI dataset with object-like masks. We can see that all components are important. We set up all models with about the same number of parameters (i.e., increase the channel number for 2D convolution and vanilla convolution) to exclude the gain from additional parameters.}
\label{tab:ablation_study}
\end{center}
\end{table}
\vspace{-5mm}



\begin{figure}
    \centering
    \includegraphics[width=\linewidth]{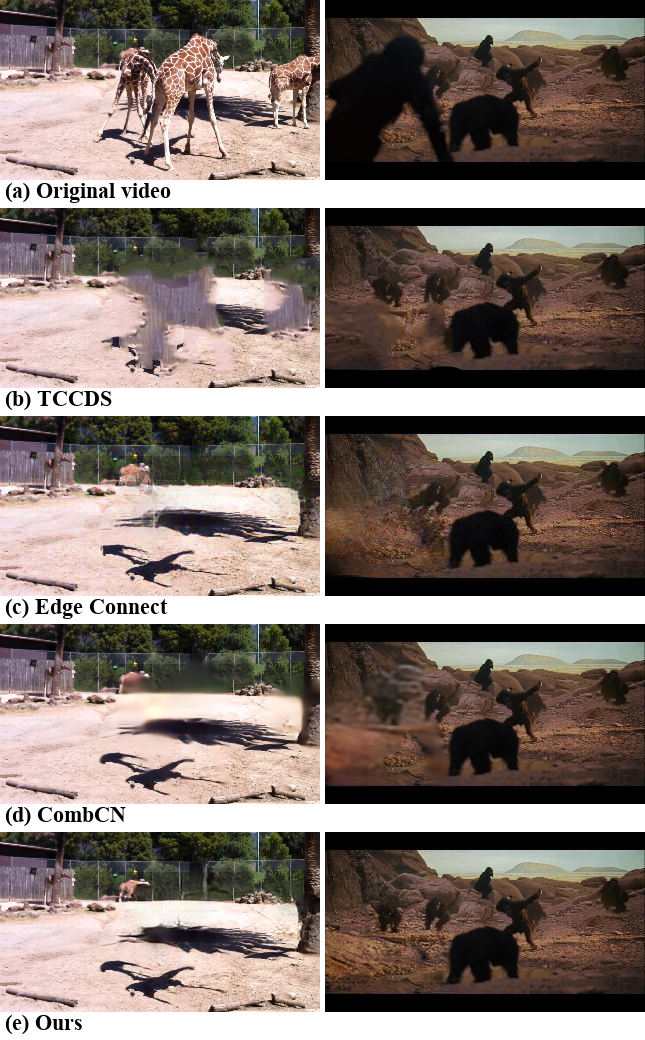}
    \caption{Our model could be easily extended to video object removal. See \href{https://bit.ly/2YcSHZ0}{videos}.}
    \label{fig:object_removal_examples}
    \vspace{-3mm}
\end{figure}

\subsection{Extension to Video Super-Resolution}
\begin{figure} [ht]
\begin{center}
\includegraphics[width=\linewidth]{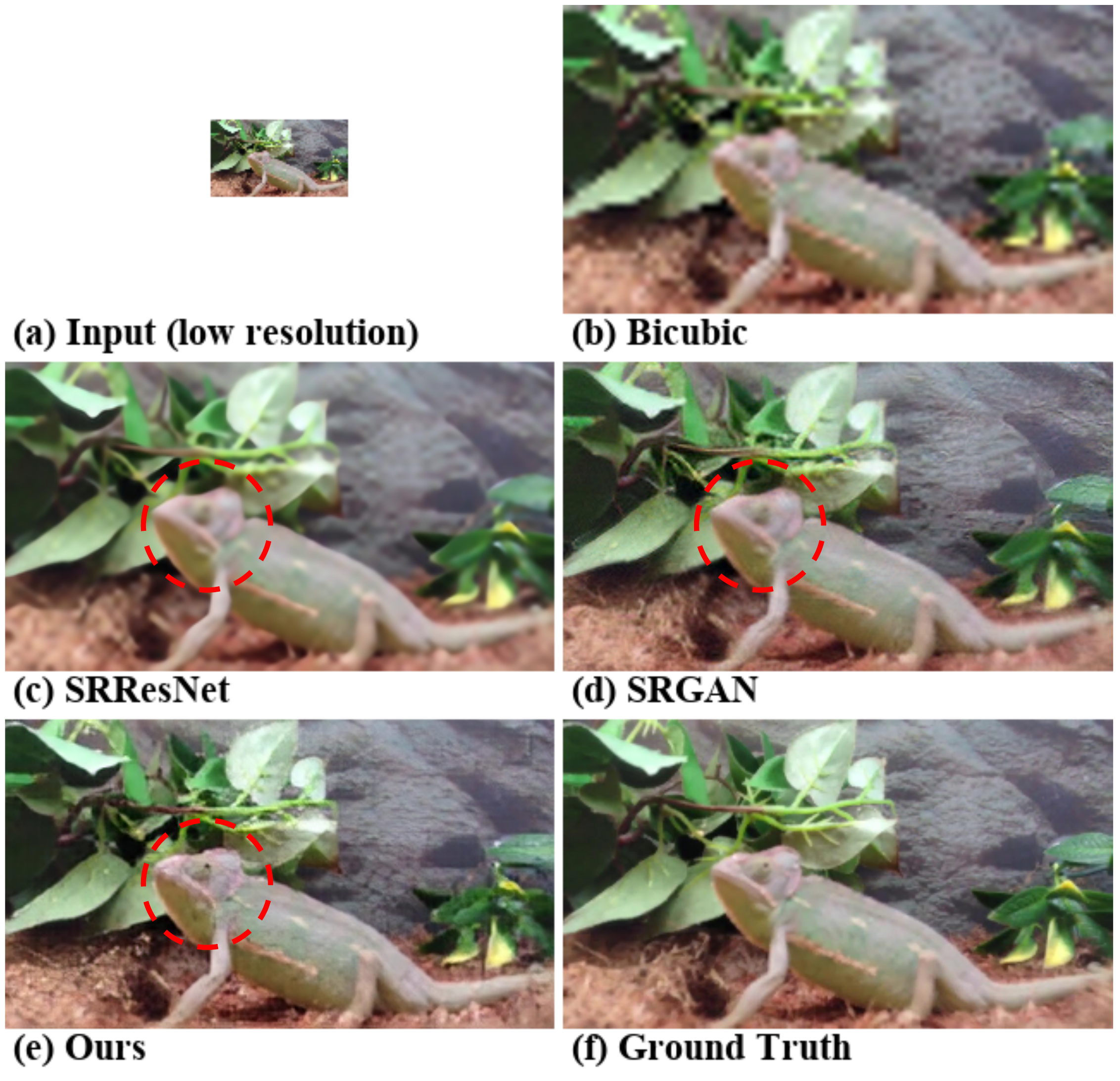}
\end{center}
\vspace{-3mm}
\caption{Two examples of spatial super-resolution with 4x up-sampling. Compared with the two baselines, our model could recover the eyes of the chameleon. 
See corresponding \href{https://bit.ly/2OijKOj}{videos}.}
\label{fig:sr_example}
\vspace{-3mm}
\end{figure}

Our model could be extended to video super-resolution, interpolation or prediction by using proper masks.
For video super-resolution, given a low-resolution video with width W, height H, length L and up-sampling factor K, we construct the input mask video in shape ($W \times K$, $H \times K$, $L$) where each pixel (x, y, t) is masked if x or y is not a multiple of K. For frame interpolation, masks could be added between frames. In Fig. \ref{fig:sr_example} and Table \ref{tab:quantitative_comparison_super_resolution}, we compare our method with well-known super-resolution methods SRResNet and SRGAN in \cite{ledig2017photo}. Our model could generate plausible high-resolution videos with low perceptual distance. See \href{https://bit.ly/2OijKOj}{result videos}.

 \begin{table}[H]
 \centering
 \begin{tabular}{|c||c|c|c|c|}
 \hline
\begin{tabular}[c]{@{}c@{}}\end{tabular} &
\begin{tabular}[c]{@{}c@{}}Bicubic \end{tabular} &
\begin{tabular}[c]{@{}c@{}}SRResNet \end{tabular} &
\begin{tabular}[c]{@{}c@{}}SRGAN \end{tabular} &
\begin{tabular}[c]{@{}c@{}}Ours \end{tabular} \\
\hline \hline 
MSE$\downarrow$ & 0.0089  & \textbf{0.0044}  &  0.0074 &  0.0076 \\
\hline 
LPIPS$\downarrow$ & 0.5141 &  0.3582 &  0.1785 &  \textbf{0.1631} \\
\hline 
FID$\downarrow$ & 1.502 &  1.083 &  \textbf{1.035} &  1.096 \\
\hline
 \end{tabular}
 \caption{Quantitative comparison for spatial super-resolution on the VOR testing set for 4x up-sample. We can see that our model could reach low perceptual quality.}
\label{tab:quantitative_comparison_super_resolution}
\vspace{-4mm}
\end{table}

\section{Discussion and Future Work}
Our model fails when the testing video is very different from the training data as most learning-based methods do. In addition, when the masked area is too thick, our model fails to generate natural results. Still, compared with the baselines, our model performs better under the two conditions (see \href{http://bit.ly/2FjTE9h}{videos}).

Besides, compared to 2D convolutions, 3D convolutions require more parameters that could lead to higher redundancy. Nonetheless, our model is single-stage, feed-forward and does not depend on optical flows, so the inference speed is fast, and the performance gain is significant. A potential solution to reduce the number of parameters is to integrate the Temporal Shift Module \cite{lin2018temporal} so that 2D convolutions could deal with temporal information. 

Also, we found that we could reach a similar performance to that of gated convolutions by simply increasing the number of channels in the ablation study. This may imply that our model still underfits the FVI dataset, or gated convolutions have less impacts for video inpainting compared to image inpainting. One of our future work would be to further compare and improve convolutional layers for free-form video inpainting.

Another future work is to integrate the user guided inputs as \cite{yu2018free, nazeri2019edgeconnect} by training the model with edge images of video frames as additional inputs. The model could generate more plausible results given the object shape information from edges. Additionally, during inference, users could draw lines to change the edge images to manipulate the output videos, which is useful for video editing. 
\vspace{-3mm}
\section{Conclusion}
In this paper, we proposed the first learning based free-form video inpainting network to our knowledge, using 3D gated convolution and a novel GAN loss. We demonstrate the power of 3D gated convolution and temporal PatchGAN to enhance video quality and temporal consistency in the video inpainting task. Our system could also be extended to video object removal, video super-resolution or video interpolation. Both the quantitative and qualitative results indicate that our model achieve state-of-the-art results. 
\vspace{-3mm}
\section{Acknowledgement}
This work was supported in part by the Ministry of Science and Technology, Taiwan, under
Grant MOST 108-2634-F-002-004. We also benefit from the NVIDIA grants and the DGX-1
AI Supercomputer. We are grateful to the National Center for High-performance Computing.

{\small
\bibliographystyle{ieee}
\bibliography{egbib}

\begin{thebibliography}{10}\itemsep=-1pt

\bibitem{abu2016youtube}
S.~Abu-El-Haija, N.~Kothari, J.~Lee, P.~Natsev, G.~Toderici, B.~Varadarajan,
  and S.~Vijayanarasimhan.
\newblock Youtube-8m: A large-scale video classification benchmark.
\newblock {\em arXiv preprint arXiv:1609.08675}, 2016.

\bibitem{barnes2009patchmatch}
C.~Barnes, E.~Shechtman, A.~Finkelstein, and D.~B. Goldman.
\newblock Patchmatch: A randomized correspondence algorithm for structural
  image editing.
\newblock In {\em ACM Transactions on Graphics (ToG)}, volume~28, page~24. ACM,
  2009.

\bibitem{bertalmio2001navier}
M.~Bertalmio, A.~L. Bertozzi, and G.~Sapiro.
\newblock Navier-stokes, fluid dynamics, and image and video inpainting.
\newblock In {\em Computer Vision and Pattern Recognition, 2001. CVPR 2001.
  Proceedings of the 2001 IEEE Computer Society Conference on}, volume~1, pages
  I--I. IEEE, 2001.

\bibitem{bertalmio2000image}
M.~Bertalmio, G.~Sapiro, V.~Caselles, and C.~Ballester.
\newblock Image inpainting.
\newblock In {\em Proceedings of the 27th annual conference on Computer
  graphics and interactive techniques}, pages 417--424. ACM
  Press/Addison-Wesley Publishing Co., 2000.

\bibitem{bornard2002missing}
R.~Bornard, E.~Lecan, L.~Laborelli, and J.-H. Chenot.
\newblock Missing data correction in still images and image sequences.
\newblock In {\em Proceedings of the tenth ACM international conference on
  Multimedia}, pages 355--361. ACM, 2002.

\bibitem{carreira2017quo}
J.~Carreira and A.~Zisserman.
\newblock Quo vadis, action recognition? a new model and the kinetics dataset.
\newblock In {\em proceedings of the IEEE Conference on Computer Vision and
  Pattern Recognition}, pages 6299--6308, 2017.

\bibitem{chang2019vornet}
Y.-L. Chang, Z.~Y. Liu, and W.~Hsu.
\newblock Vornet: Spatio-temporally consistent video inpainting for object
  removal.
\newblock {\em arXiv preprint arXiv:1904.06726}, 2019.

\bibitem{drori2003fragment}
I.~Drori, D.~Cohen-Or, and H.~Yeshurun.
\newblock Fragment-based image completion.
\newblock In {\em ACM Transactions on graphics (TOG)}, volume~22, pages
  303--312. ACM, 2003.

\bibitem{gatys2015neural}
L.~A. Gatys, A.~S. Ecker, and M.~Bethge.
\newblock A neural algorithm of artistic style.
\newblock {\em arXiv preprint arXiv:1508.06576}, 2015.

\bibitem{goodfellow2014generative}
I.~Goodfellow, J.~Pouget-Abadie, M.~Mirza, B.~Xu, D.~Warde-Farley, S.~Ozair,
  A.~Courville, and Y.~Bengio.
\newblock Generative adversarial nets.
\newblock In {\em Advances in neural information processing systems}, pages
  2672--2680, 2014.

\bibitem{granados2012not}
M.~Granados, J.~Tompkin, K.~Kim, O.~Grau, J.~Kautz, and C.~Theobalt.
\newblock How not to be seen—object removal from videos of crowded scenes.
\newblock In {\em Computer Graphics Forum}, volume~31, pages 219--228. Wiley
  Online Library, 2012.

\bibitem{heusel2017gans}
M.~Heusel, H.~Ramsauer, T.~Unterthiner, B.~Nessler, and S.~Hochreiter.
\newblock Gans trained by a two time-scale update rule converge to a local nash
  equilibrium.
\newblock In {\em Advances in Neural Information Processing Systems}, pages
  6626--6637, 2017.

\bibitem{Huang-SigAsia-2016}
J.-B. Huang, S.~B. Kang, N.~Ahuja, and J.~Kopf.
\newblock Temporally coherent completion of dynamic video.
\newblock {\em ACM Transactions on Graphics (TOG)}, 35(6):196, 2016.

\bibitem{iizuka2017globally}
S.~Iizuka, E.~Simo-Serra, and H.~Ishikawa.
\newblock Globally and locally consistent image completion.
\newblock {\em ACM Transactions on Graphics (ToG)}, 36(4):107, 2017.

\bibitem{johnson2016perceptual}
J.~Johnson, A.~Alahi, and L.~Fei-Fei.
\newblock Perceptual losses for real-time style transfer and super-resolution.
\newblock In {\em European conference on computer vision}, pages 694--711.
  Springer, 2016.

\bibitem{ledig2017photo}
C.~Ledig, L.~Theis, F.~Husz{\'a}r, J.~Caballero, A.~Cunningham, A.~Acosta,
  A.~Aitken, A.~Tejani, J.~Totz, Z.~Wang, et~al.
\newblock Photo-realistic single image super-resolution using a generative
  adversarial network.
\newblock In {\em Proceedings of the IEEE conference on computer vision and
  pattern recognition}, pages 4681--4690, 2017.

\bibitem{lin2018temporal}
J.~Lin, C.~Gan, and S.~Han.
\newblock Temporal shift module for efficient video understanding.
\newblock {\em arXiv preprint arXiv:1811.08383}, 2018.

\bibitem{liu2018image}
G.~Liu, F.~A. Reda, K.~J. Shih, T.-C. Wang, A.~Tao, and B.~Catanzaro.
\newblock Image inpainting for irregular holes using partial convolutions.
\newblock {\em arXiv preprint arXiv:1804.07723}, 2018.

\bibitem{miyato2018spectral}
T.~Miyato, T.~Kataoka, M.~Koyama, and Y.~Yoshida.
\newblock Spectral normalization for generative adversarial networks.
\newblock {\em arXiv preprint arXiv:1802.05957}, 2018.

\bibitem{nazeri2019edgeconnect}
K.~Nazeri, E.~Ng, T.~Joseph, F.~Qureshi, and M.~Ebrahimi.
\newblock Edgeconnect: Generative image inpainting with adversarial edge
  learning.
\newblock 2019.

\bibitem{newson2014video}
A.~Newson, A.~Almansa, M.~Fradet, Y.~Gousseau, and P.~P{\'e}rez.
\newblock Video inpainting of complex scenes.
\newblock {\em SIAM Journal on Imaging Sciences}, 7(4):1993--2019, 2014.

\bibitem{pathak2016context}
D.~Pathak, P.~Krahenbuhl, J.~Donahue, T.~Darrell, and A.~A. Efros.
\newblock Context encoders: Feature learning by inpainting.
\newblock In {\em Proceedings of the IEEE Conference on Computer Vision and
  Pattern Recognition}, pages 2536--2544, 2016.

\bibitem{real2017youtube}
E.~Real, J.~Shlens, S.~Mazzocchi, X.~Pan, and V.~Vanhoucke.
\newblock Youtube-boundingboxes: A large high-precision human-annotated data
  set for object detection in video.
\newblock In {\em Proceedings of the IEEE Conference on Computer Vision and
  Pattern Recognition}, pages 5296--5305, 2017.

\bibitem{rossler2018faceforensics}
A.~R{\"o}ssler, D.~Cozzolino, L.~Verdoliva, C.~Riess, J.~Thies, and
  M.~Nie{\ss}ner.
\newblock Faceforensics: A large-scale video dataset for forgery detection in
  human faces.
\newblock {\em arXiv preprint arXiv:1803.09179}, 2018.

\bibitem{russakovsky2015imagenet}
O.~Russakovsky, J.~Deng, H.~Su, J.~Krause, S.~Satheesh, S.~Ma, Z.~Huang,
  A.~Karpathy, A.~Khosla, M.~Bernstein, et~al.
\newblock Imagenet large scale visual recognition challenge.
\newblock {\em International journal of computer vision}, 115(3):211--252,
  2015.

\bibitem{simonyan2014very}
K.~Simonyan and A.~Zisserman.
\newblock Very deep convolutional networks for large-scale image recognition.
\newblock {\em arXiv preprint arXiv:1409.1556}, 2014.

\bibitem{wang2018videoinp}
C.~Wang, H.~Huang, X.~Han, and J.~Wang.
\newblock Video inpainting by jointly learning temporal structure and spatial
  details.
\newblock In {\em Proceedings of the 33th AAAI Conference on Artificial
  Intelligence}, 2019.

\bibitem{wang2018video}
T.-C. Wang, M.-Y. Liu, J.-Y. Zhu, G.~Liu, A.~Tao, J.~Kautz, and B.~Catanzaro.
\newblock Video-to-video synthesis.
\newblock {\em arXiv preprint arXiv:1808.06601}, 2018.

\bibitem{wexler2007space}
Y.~Wexler, E.~Shechtman, and M.~Irani.
\newblock Space-time completion of video.
\newblock {\em IEEE Transactions on Pattern Analysis \& Machine Intelligence},
  (3):463--476, 2007.

\bibitem{xie2012image}
J.~Xie, L.~Xu, and E.~Chen.
\newblock Image denoising and inpainting with deep neural networks.
\newblock In {\em Advances in neural information processing systems}, pages
  341--349, 2012.

\bibitem{xu2018youtube}
N.~Xu, L.~Yang, Y.~Fan, J.~Yang, D.~Yue, Y.~Liang, B.~Price, S.~Cohen, and
  T.~Huang.
\newblock Youtube-vos: Sequence-to-sequence video object segmentation.
\newblock In {\em Proceedings of the European Conference on Computer Vision
  (ECCV)}, pages 585--601, 2018.

\bibitem{yan2018shift}
Z.~Yan, X.~Li, M.~Li, W.~Zuo, and S.~Shan.
\newblock Shift-net: Image inpainting via deep feature rearrangement.
\newblock {\em arXiv preprint arXiv:1801.09392}, 2018.

\bibitem{yu2018free}
J.~Yu, Z.~Lin, J.~Yang, X.~Shen, X.~Lu, and T.~S. Huang.
\newblock Free-form image inpainting with gated convolution.
\newblock {\em arXiv preprint arXiv:1806.03589}, 2018.

\bibitem{yu2018generative}
J.~Yu, Z.~Lin, J.~Yang, X.~Shen, X.~Lu, and T.~S. Huang.
\newblock Generative image inpainting with contextual attention.
\newblock {\em arXiv preprint}, 2018.

\bibitem{zhang2018unreasonable}
R.~Zhang, P.~Isola, A.~A. Efros, E.~Shechtman, and O.~Wang.
\newblock The unreasonable effectiveness of deep features as a perceptual
  metric.
\newblock {\em arXiv preprint}, 2018.

\bibitem{zhou2017places}
B.~Zhou, A.~Lapedriza, A.~Khosla, A.~Oliva, and A.~Torralba.
\newblock Places: A 10 million image database for scene recognition.
\newblock {\em IEEE Transactions on Pattern Analysis and Machine Intelligence},
  2017.

\end{thebibliography}
}

\end{document}